\documentclass[conference]{IEEEtran}
\IEEEoverridecommandlockouts
\usepackage{cite}
\usepackage{amsmath,amssymb,amsfonts}
\usepackage{algorithmic}

\usepackage{graphicx}
\usepackage{textcomp}
\usepackage{xcolor}
\graphicspath{ {Images/} }
\def\BibTeX{{\rm B\kern-.05em{\sc i\kern-.025em b}\kern-.08em
    T\kern-.1667em\lower.7ex\hbox{E}\kern-.125emX}}
\begin{document}
\title{Playing a 2D Game Indefinitely using NEAT and Reinforcement Learning\\
}
\author{\IEEEauthorblockN{Jerin Paul Selvan}
\IEEEauthorblockA{\textit{Dept. of Computer Engineering} \\
\textit{Pune Institute of Computer Technology}\\
Pune, India \\
jerinsprograms@gmail.com}
\and
\IEEEauthorblockN{Dr. P. S. Game}
\IEEEauthorblockA{\textit{Dept. of Computer Engineering} \\
\textit{Pune Institute of Computer Technology}\\
Pune, India \\
psgame@pict.edu}
}
\maketitle
\begin{abstract}
For over a decade now, robotics and the use of artificial agents have become a common thing. Testing the performance of new path finding or search space optimisation algorithms has also become a challenge as they require simulation or an environment to test them. The creation of artificial environments with artificial agents is one of the methods employed to test such algorithms. Games have also become an environment to test them. The performance of the algorithms can be compared by using artificial agents that will behave according to the algorithm in the environment they are put in. The performance parameters can be, how quickly the agent is able to differentiate between rewarding actions and hostile actions. This can be tested by placing the agent in an environment with different types of hurdles and the goal of the agent is to reach the farthest by taking decisions on actions that will lead to avoiding all the obstacles. The environment chosen is a game called "Flappy Bird". The goal of the game is to make the bird fly through a set of pipes of random heights. The bird must go in between these pipes and must not hit the top, the bottom, or the pipes themselves. The actions that the bird can take are either to flap its wings or drop down with gravity. The algorithms that are enforced on the artificial agents are NeuroEvolution of Augmenting Topologies (NEAT) and Reinforcement Learning. The NEAT algorithm takes an ’N’ initial population of artificial agents. They follow genetic algorithms by considering an objective function, crossover, mutation, and augmenting topologies. Reinforcement learning, on the other hand, remembers the state, the action taken at that state, and the reward received for the action taken using a single agent and a Deep Q-learning Network. The performance of the NEAT algorithm improves as the initial population of the artificial agents is increased.\\
\end{abstract}
\section{Introduction}
An intelligent agent is anything that can detect its surroundings, act independently to accomplish goals, and learn from experience or use knowledge to execute tasks better. The agent’s surroundings are considered an environment in artificial intelligence. The agent uses actuators to send its output to the environment after receiving information from it through sensors. \cite{b11} There are several types of environments, Fully Observable vs Partially Observable, Deterministic vs Stochastic, Competitive vs Collaborative, Single-agent vs Multi-agent, Static vs Dynamic, Discrete vs Continuous, Episodic vs Sequential and Known vs Unknown. An approach to machine learning known as NEAT, or Neuroevolution of Augmenting Topologies, functions similarly to evolution. In its most basic form, \cite{b1} NEAT is a technique for creating networks that are capable of performing a certain activity, like balancing a pole or operating a robot. It’s significant that NEAT networks can learn using a reward function as opposed to back-propagation. By executing actions and observing the outcomes of those actions, an agent learns how to behave in a given environment via reinforcement learning, a feedback-based machine learning technique. The agent receives compliments for each positive activity and is penalised or given negative feedback for each negative action. In contrast to supervised learning, reinforcement learning uses feedback to autonomously train the agent without the use of labelled data. The agent can only learn from its experience because there is no labelled data. In situations like gaming, robotics, and the like, where decisions must be made sequentially and with a long-term objective, RL provides a solution. The agent engages with the environment and independently explores it. In reinforcement learning, an agent’s main objective is to maximise positive rewards while doing better. 
\section{Literature Survey}
Games have been used a lot to act as an environment to test algorithms. There is a lot of research \cite{b3} done to create an AI bot that can challenge a player in a multi-player or two-player game. Neuroevolution and Reinforcement learning algorithms are some of the algorithms that are used to create AI bots or artificial agents. \cite{b1}, \cite{b7} and \cite{b8} have implemented a configuration of an ANN called Neuroevolution. The algorithm does not depend on the actions taken by the agents as a whole. \cite{b3}, \cite{b4}, \cite{b5}, \cite{b6} and \cite{b7} use Reinforcement Learning algorithm with Deep Q-Learning to train the agents.\\ 
The performance of the Neuroevolution algorithm depends on the objective function, initial population, mutation rate, weights and bias added to the network, the activation function used and overall topology of the network. Authors in \cite{b2} talk about how superior the Neuroevolution algorithm is over the traditional Reinforcement Learning algorithm with the Deep Q-Learning algorithm. Neuroevolution has an upper hand when it comes to the time taken by the artificial agent to train itself. There are other parameters that need to be taken into consideration while using a Neural Network. The topology of the network plays a vital role in the performance. Two strategies were proposed by Evgenia Papavasileiou (2021) \cite{b2}, using fixed topologies in the neural networks and using augmented topologies. The network topology is a single hidden layer of neurons, with each hidden neuron connected to every network input and every network output. Evolution searches the space of connection weights of this fully-connected topology by allowing high performing networks to reproduce. The weight space is explored through the crossover of network weight vectors and through the mutation of single networks’ weights. Thus, the goal of fixed-topology NE is to optimise the connection weights that determine the functionality of a network. The topology, or structure, of neural networks also affects their functionality, and modifying the network structure has been effective as part of supervised training.\\
There are two ways of making use of the environment. Authors in \cite{b3}, \cite{b4}, \cite{b6} and \cite{b7} use DNN to extract the features from the frame of the game and they form the input to the agent. However, \cite{b1}, \cite{b5} and \cite{b8} make use of the game itself and place the agent to perceive its surroundings. There are several combinations of Reinforcement Learning algorithms possible, like Deep Neural Networks (DNN), Long short-term memory (LSTM), Deep Q-Network (DQN) and the like. However, depending on the type of obstacle and the type of game, its performance varies.\\
Reinforcement Learning algorithm with DNN and LSTM have been used in \cite{b3}. This algorithm addresses issues like vast search space, dependencies between the actions taken by the agent, the state and the environment, inputs and imperfect information. To reduce the complexity of the data generated by the perception of the agent, data skipping techniques are implemented. There is, however, a drawback with this algorithm. It takes a lot of time for the agent to train. Or, for every discrete step taken by the agent, it receives a state that belongs to a set S and it sends an action from the set A actions to the environment. The environment makes a transition from state S\textsubscript{t} to S\textsubscript{t+1} and a gamma value [0, 1] determines the preference for immediate reward over long-term reward. A self-playing method is used by storing the parameters of the network to create a pool of past agents. This pool of past agents is used to sample opponents. This method offers RL to learn the Nash equilibrium strategy. Data skipping techniques were proposed in this paper. It refers to the process of dropping certain data during the training and evaluation process. Data skipping techniques proposed are: "no-op" and "maintain move decision". The network is composed of an LSTM-based architecture, which has four heads with a shared state representation layer. An actor-critic off-policy learning algorithm was proposed.\\
Botong Liu (2020) \cite{b4} has used Reinforcement Learning with DQN. The game was split into frames, and each game image was sequentially scaled, grayed, and adjusted for brightness. Deep Q Network algorithm was used to convert the game decision problem into a classification and recognition problem of multi-dimensional images and solve it using CNN. Reinforcement learning works best for continuous decision-making problems. However, Deep Reinforcement Learning has a limitation of not converging for which Neural fitted Q-learning and DQN algorithms were used to overcome the issue. Since FNQ can work with numerical information only the author suggests use of DQN. Combining Q learning with CNN, the DQN can achieve self-learning. ReLu and maximum pooling layers are added to the CNN. Gradient descent (Adam Optimizer) was used to train the DQN parameters.\\
Q-Value function based algorithms are the focus of Aidar Shakerimov (2021) \cite{b5}. For the DQN algorithms, improvements could be achieved in their performance by using a cumulative reward for training actions. To speed up training, RNN-ReLU was used instead of LSTM or GRU. LSTM or GRU performs better than RNN-ReLU but takes 7 times more time to train. Label smoothing was used to prevent the vanishing gradients in RNN-ReLU. However, DQN is sensitive to seed randomization.\\
SARSA is a slight variation of the traditional Q-Learning algorithm. Authors in \cite{b6} use SARSA and Q-Learning algorithms with modifications such as $\epsilon$-greedy policy, discretization and backward updates. Some variants of Q-Learning were also implemented such as a tabular approach, Q-value approximation using linear regression, and NN. In the implementation, \cite{b6} finds the SARSA algorithm to have outperformed Q-learning. The specifications of the rewards are a positive 5 for passing a pipe, a negative 1000 for hitting a pipe, and a positive 0.5 for surviving a frame. Feed-forward NN was used with a 3 neuron input layer, 50 neuron hidden layer, 20 neuron hyphen layer, and a 2 neuron output layer (ReLU activation function). The CNN is used with preprocessed input image by removing the background, grayscale, and resizing to 80 x 80, 2 CNN layers were used, one using sixteen 5 × 5 kernels with stride 2, and another with thirty-two 5 × 5 kernels with stride 2. \\
\cite{b7} proposes the use of specific feature selection and presents the state by the bird velocity and the difference between the bird’s position and the next lower pipe. This reduces the feature space and eliminates the need for deeper modules. The agent is provided with rational human-level inputs along with generic RL and a standard 3-layer NN with a genetic optimization algorithm. The reward for the agent is a positive 1 for every cross of the pipe and a negative 100 if the agent dies. The Neuro evolution has the following characteristics: the NN weights and the number of hidden layer units undergo changes, the mutation rate is kept at 0.3, and the initial population size is 200. 
\cite{b8} proposes the use of two levels for the Flappy Bird game. The fitness function is calculated by the distance traveled by the agent and the current distance to the closest gap. The mutation rate is kept at 0.2, and there are 5 neurons in the hidden layer.
\section{Methodology}
The NEAT algorithm implementation is dependent on the objective function, crossover, mutation, and a population of agents. For a given position of the bird, say (x, y), there are two actions that the agent can make. Either the bird flaps its wings or it does not flap its wings. The vertical and horizontal distances traveled by the agent are determined by the following equations.
\begin{equation}
d_{vertical} = v_{jump}.t + \frac{1}{2}.a.t^2\label{eq1}
\end{equation}
\begin{equation}
d_{horizontal} = v_{floor}.t\label{eq2}
\end{equation}
\begin{equation}
d_{floor} = v_{floor}.t\label{eq3}
\end{equation}
\begin{equation}
d_{pipe} = v_{pipe}.t\label{eq4}
\end{equation}
Eq.~\eqref{eq1} determines the vertical displacement of the agent, where \emph{a} is the acceleration that is a constant \cite{b12}.
\begin{figure}[htbp]
\centerline{\includegraphics[scale=0.4]{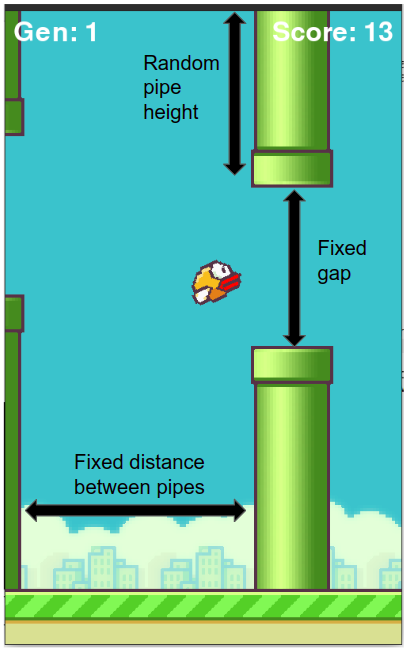}}
\caption{Details of the game environment \cite{b10}}
\label{fig1}
\end{figure}
As shown in the Fig.~\ref{fig2}, the y coordinate of the agent, the distance between the top pipe and the agent \emph{(y - T')} and the distance between the bottom pipe and the agent \emph{(T')} are the inputs to the neural network. The gap between the top and the bottom pipe is fixed to 320 pixels, and the heights are randomly generated. The distance between subsequent pipes is also kept constant. With respect to the NEAT algorithm, the fitness of the agent is determined by the number of pipes that the agent is able to cross without collision. As soon as the agent collides with the pipe, hits the roof, or falls down to the ground, the agent is removed from the environment. The performance of the algorithm depends on the initial population that is taken into consideration. The activation function used is the hyperbolic tangent function. The mutation rate is kept at 0.03.
\begin{figure}[htbp]
\centerline{\includegraphics[scale=0.386]{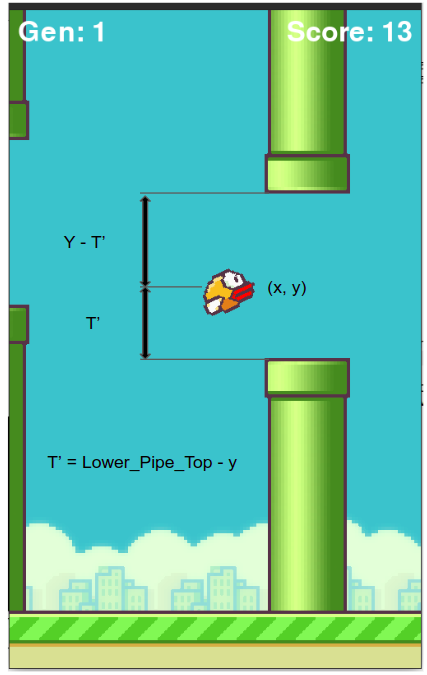}}
\caption{Parameters required as input to the NN}
\label{fig2}
\end{figure}
The encoding of the chromosome is shown in Table. \ref{tab1}. The weight of the connection from a node in a layer to another node in the other layer and the dropped value is also considered as part of the encoding. If the connection is to be dropped, it is encoded with the value 0 otherwise, it has the value 1. 
\begin{figure}[htbp]
\centerline{\includegraphics[scale=0.4]{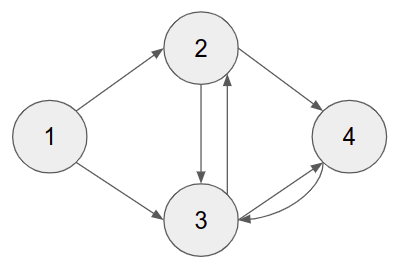}}
\caption{Diagramatic view of the encoded chromosome in Table \ref{tab1}}
\label{fig3}
\end{figure}
\begin{table}[htbp]
\caption{Encoding of a chromosome before crossover and mutation}
\begin{center}
\begin{tabular}{|c|c|c|c|c|c|c|c|}
\hline
\textbf{Weight} & 0.25 & 2.31 & 1.55 & 0.98 & 5.11 & 1.17 & 0.07 \\
\hline
\textbf{From} & 1 & 2 & 3 & 1 & 3 & 4 & 2 \\
\hline
\textbf{To} & 2 & 3 & 2 & 3 & 4 & 3 & 4 \\
\hline
\textbf{Enabled} & 1 & 0 & 1 & 1 & 1 & 1 & 1 \\
\hline
\end{tabular}
\label{tab1}
\end{center}
\end{table}
With reference to Fig. \ref{fig3} and Table. \ref{tab1}, the edges between the nodes are represented by the rows 'From' and 'To'. The Table. \ref{tab1} shows the encoding of the network before mutation. After the mutation, or rather after topology augmentation, the encoding of the edges is shown in Table. \ref{tab2}. The resultant connections are shown in Fig. \ref{fig4}. The edges that are in red are the edges that were dropped, and the edges that are in green are the ones that have been added as a result of the mutation. The cross-over process happens between any two randomly selected parents. The next population is determined by the fitness of the individual agents.
\begin{figure}[htbp]
\centerline{\includegraphics[scale=0.4]{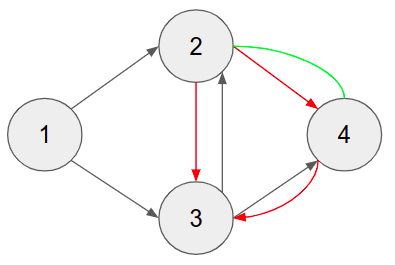}}
\caption{Diagramatic view of the encoded chromosome in Table. \ref{tab2}}
\label{fig4}
\end{figure}
\begin{table}[htbp]
\caption{Encoding of a chromosome after crossover and mutation}
\begin{center}
\begin{tabular}{|c|c|c|c|c|c|c|c|}
\hline
\textbf{Weight} & 0.25 & 5.11 & 1.17 & 0.98 & 2.31 & 1.55 & 0.07 \\
\hline
\textbf{From} & 1 & 2 & 4 & 1 & 3 & 3 & 4 \\
\hline
\textbf{To} & 3 & 4 & 2 & 4 & 2 & 4 & 3 \\
\hline
\textbf{Enabled} & 1 & 1 & 1 & 0 & 1 & 1 & 0 \\
\hline
\end{tabular}
\label{tab2}
\end{center}
\end{table}
\section{Results}
\begin{figure}[htbp]
\centerline{\includegraphics[scale=0.5]{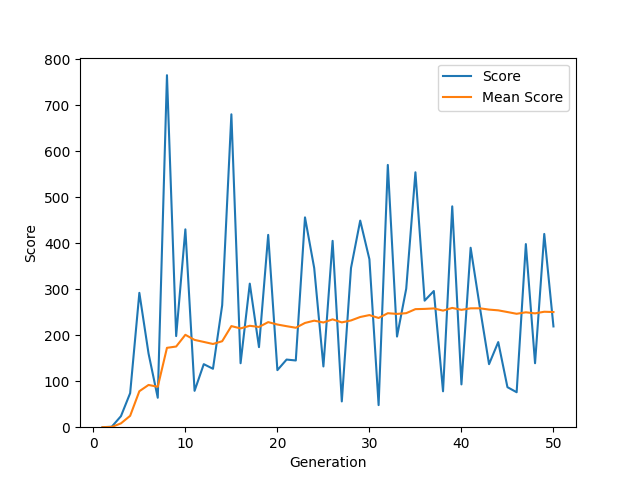}}
\caption{Gameplay when initial population is 80}
\label{fig5}
\end{figure}

\begin{figure}[htbp]
\centerline{\includegraphics[scale=0.5]{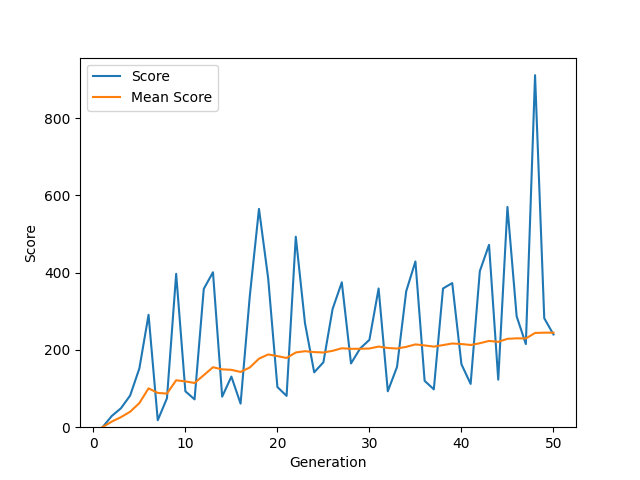}}
\caption{Gameplay when initial population is 100}
\label{fig6}
\end{figure}

\begin{figure}[htbp]
\centerline{\includegraphics[scale=0.5]{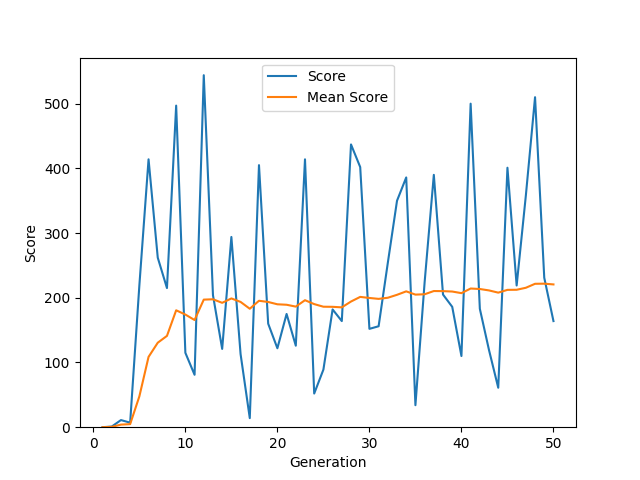}}
\caption{Gameplay when initial population is 120}
\label{fig7}
\end{figure}

\begin{figure}[htbp]
\centerline{\includegraphics[scale=0.5]{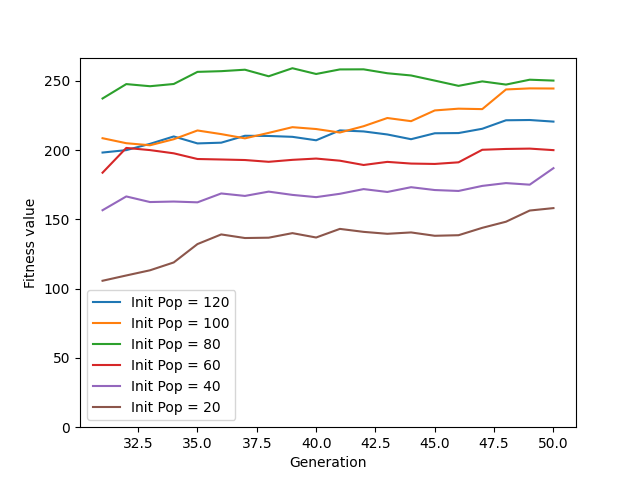}}
\caption{Average scores over initial population change (Gen 30 - Gen 50)}
\label{fig8}
\end{figure}

\begin{figure}[htbp]
\centerline{\includegraphics[scale=0.5]{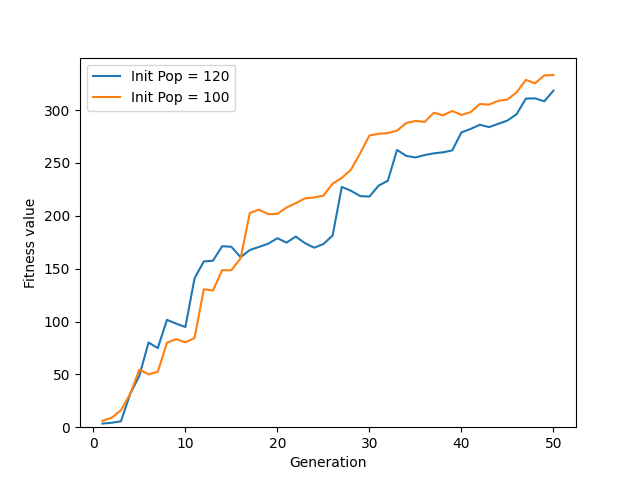}}
\caption{Average Fitness of the population over initial population change}
\label{fig9}
\end{figure}

\begin{figure}[htbp]
\centerline{\includegraphics[scale=0.5]{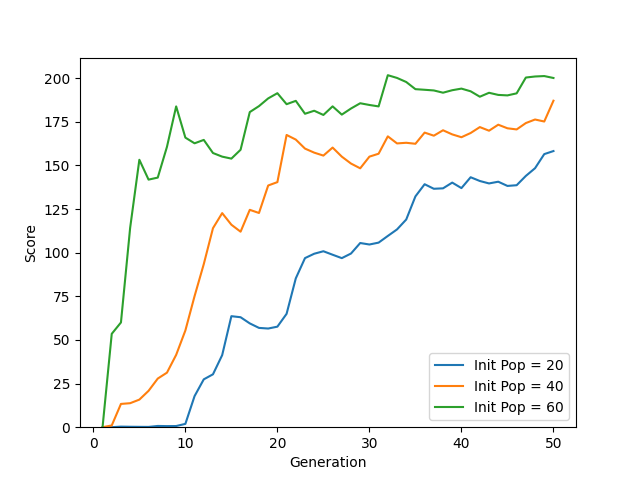}}
\caption{Speed of agents getting trained over initial population change}
\label{fig10}
\end{figure}

The implementation of the algorithm requires no historic data or any dataset. The algorithm makes use of the sensory data perceived from the environment by the artificial agent as the program runs. The inputs to the algorithm are the \textbf{\emph{y}} position of the agent, the vertical distance of the agent from the top pipe, and the vertical distance of the agent from the lower pipe. The output of the algorithm is the action that the agent is to take i.e. jump or drop down owing to gravity. NEAT algorithm was implemented by taking different initial populations. Fig. \ref{fig5}, Fig. \ref{fig6} and Fig. \ref{fig7} shows the average score and the scores reached in every generation, when the game is played by the agents over 50 generations.
The change in the average scores over the change in the initial population is separately shown in Fig. \ref{fig8} for generations 30 to 50. The average score of the agent is steadily increasing from when the initial population is 20 to 100. The maximum score is observed when the population is 160. The average fitness value of the population is higher when the initial population size is 100. This is shown in Fig. \ref{fig9}. The initial training phase is less than 5 generations. When the initial population has fewer agents, it takes more generations to spike the average score of the game. This can be observed from Fig. \ref{fig10}.
Table. \ref{tab3} shows the average score and the maximum score gained by the agent over 50 generations. A maximum score of 1025 is obtained when the initial population is 160 and the gameplay run till 50 generations.
\begin{table}[htbp]
\caption{Scores over change in initial population}
\begin{center}
\begin{tabular}{|c|c|c|}
\hline
\textbf{Initial Population} & \textbf{Average Score} & \textbf{Max Score}\\
\hline
20 & 158.2 & 583 \\
\hline
40 & 187.04 & 771 \\
\hline
60 & 200.06 & 756 \\
\hline
80 & 250.28 & 765 \\
\hline
100 & 244.56 & 911 \\
\hline
120 & 220.66 & 544 \\
\hline
140 & 255.82 & 565 \\
\hline
160 & 293.72 & 1025 \\
\hline
\end{tabular}
\label{tab3}
\end{center}
\end{table}
\section*{Conclusion and Future Scope}
By using a 2D game, the performance of the algorithms can be determined very efficiently. Unlike simulation, the creation of an environment gives better control over the environment. Through various iterations by changing the initial population size, the average score gained by the agent has increased. The initial population of agents also affects the training speed. The more the agents, the quicker the training is done. The highest average score is obtained when the initial population is set to 100 individuals. It can be concluded that the performance of the algorithm increases as the initial population is increased. The implemented algorithm can be extended to making use of Reinforcement Learning with multiple agents and using Augmented Topologies along with the Deep Q-Learning model.
         
\end{document}